С.В. Поликарпов, В.С. Дергачёв, К.Е. Румянцев, Д.М. Голубчиков

# Новая модель искусственного нейрона: кибернейрон и области его применения

**Аннотация.** В данной статье рассматривается новый тип искусственного нейрона, названный авторами «кибернейроном». В отличие от классических моделей искусственных нейронов данный тип нейрона использует табличные подстановки вместо операции умножения входных значений на весовые коэффициенты. Это позволило значительно увеличить информационную ёмкость отдельного нейрона, а также значительно упростить процедуру обучения. Приводится пример использования кибернейрона при решении задачи обнаружения компьютерных вирусов.

**Ключевые слова:** кибернейрон, нейросеть, обнаружение компьютерных вирусов.

Спектр задач, решаемых искусственными нейронными сетями (нейросетями) достаточно широк. Это объясняется основными свойствами, присущих нейросетям: возможность обучения нейросети в процессе её работы; способность обобщать получаемую информацию; высокая отказоустойчивость; параллельность вычислений и т.д.

Необходимо отметить, что искусственные нейросети являются очень грубой математической моделью природных нейросетей. При этом существует два подхода к построению искусственных нейросетей:

**1. Первый подход** [1-3] предполагает, что сам нейрон является простейшим вычислительным элементом, а решение сложных задач достигается использованием сложной конфигурации нейросети. Это позволило создать достаточно простую модель искусственного нейрона (формальный нейрон). Что в своё время дало сильный толчок в создании различных моделей искусственных нейронных сетей (персептрон Розенблатта, сети Хопфилда, самоорганизующиеся карты Кохонена и т.д.). Именно модель формального нейрона используется в настоящее время при решении практических задач распознавания образов, классификации, прогнозирования и т.д.

Недостатки:

– нет чётких критериев (правил) выбора количества нейронов и связей между ними. Составление искусственной нейронной сети, способной решать конкретную поставленную задачу, является достаточно трудоёмким процессом и является своеобразным искусством;

– относительно небольшая информационная ёмкость отдельного искусственного нейрона (т.е. количество запоминаемых нейроном образов). Чем больше нейросеть должна «помнить» образов, тем более сложной должна быть нейросеть (и количество входящих в неё нейронов). В купе с предыдущим недостатком это значительно затрудняет составления искусственной нейросети, предназначенной для решения сложных задач;

– сложность (ресурсоёмкость) реализации искусственного нейрона в аппаратном виде. Объясняется наличием операций умножения на каждом из входов нейрона.

Отличие от реальных нейронов:

– не учитывается пластичность весов во времени;

– алгоритм обучения на основе обратного распространения ошибки не имеет аналога в природных нейросетях;

– остальные методы обучения искусственных нейросетей являются вычислительно неэффективными;

**2. Второй подход** предполагает, что отдельный нейрон является крайне сложным вычислительным элементом [3, 4]. В отличие от первого подхода, в котором предполагается, что нейрон является простейшим вычислительным элементом, в данном подходе предполагается, что нейрон сам по себе является сложным вычислительным элементом. Например, в [3] обосновывается, что только количество информации в одном нейроне может достигать $10^{14}$ бит. Последователи этого подхода пытаются создать точную математическую модель отдельного нейрона и всей нейросети в целом. Существуют достаточно сложные математические модели природных нейронов, однако достоверно неизвестно, насколько эти модели соответствуют реальности. Также нет свидетельств использования этих моделей при решении какой-либо практической задачи.

**Рассмотрим биологический нейрон [5,6].** Нейрон (нервная клетка) является особой биологической клеткой, которая обрабатывает информацию (рисунок 1). Она состоит из тела клетки (cell body), или сомы (soma), и двух типов внешних древоподобных ветвей: аксона (axon) и дендритов (dendrites). Нейрон получает сигналы (импульсы) от других нейронов через дендриты (приемники) и передает сигналы, сгенерированные телом клетки, вдоль аксона (передатчик), который в конце разветвляется на волокна (strands). На окончаниях этих волокон находятся синапсы (synapses).

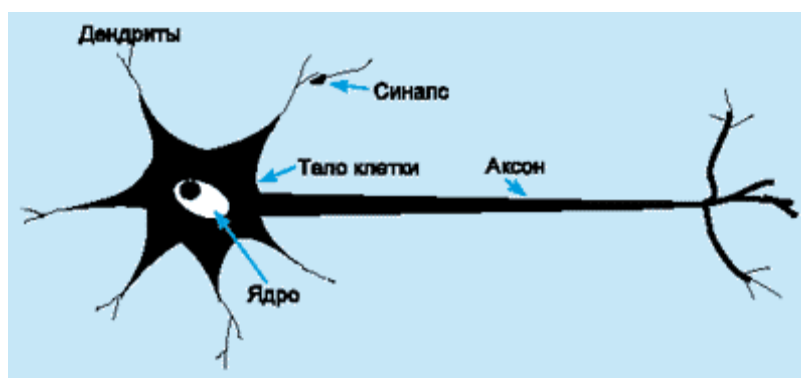

***Рисунок 1.*** *Схема биологического нейрона*

Синапс (рисунок 2) является элементарной структурой и функциональным узлом между двумя нейронами (волокно аксона одного нейрона и ден-дрит другого). Когда импульс достигает синаптического окончания, высвобождаются определенные химические вещества, называемые нейротрансмиттерами (медиаторами). Медиаторы диффундируют через синаптическую щель и попадают на рецепторы нейрона-приёмника, возбуждая или затормаживая способность нейрона-приемника генерировать электрические импульсы. Результативность синапса может настраиваться проходящими через него сигналами, так что синапсы могут обучаться в зависимости от активности процессов, в которых они участвуют. Эта зависимость от предыстории действует как память.

Нейроны взаимодействуют посредством короткой серии импульсов, как правило, продолжительностью несколько мсек. Сообщение передается посредством частотно-импульсной модуляции. Частота может изменяться от нескольких единиц до сотен герц. Сложные решения по восприятию информации, как, например, распознавание лица, человек принимает за несколько сотен мс. Эти решения контролируются сетью нейронов, которые имеют скорость выполнения операций всего несколько мс. Это означает, что вычисления требуют не более 100 последовательных стадий. Другими словами, для таких сложных задач мозг "запускает" параллельные программы, содержащие около 100 шагов. Это известно как правило ста шагов [5,6].

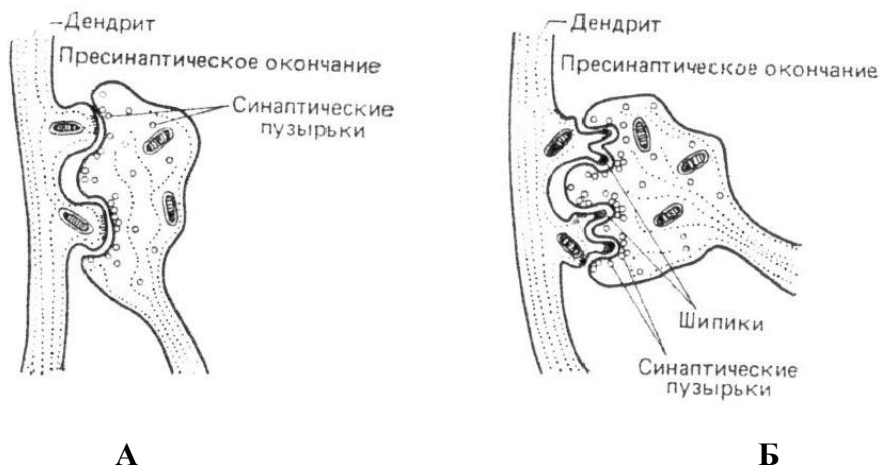

**А**            **Б**

*Рисунок 2. Пример синапсов нейронов головного мозга собаки [12]: а – животное, выращенное в темноте; б – нормальное животное*

**Предлагаемый подход:**

Если проанализировать известные данные, то можно выделить следующие противоречащие факты:

1. Синапсы природных нейронов, которые чаще активируются, имеют больший размер (более развитую структуру). Считается, что каждый синапс нейрона работает как возбуждающий или подавляющий элемент, и что увеличение размеров синапса приводит к увеличению его весового коэффициента.

2. Синапс состоит из большого количества рецепторов различного типа, которые по-разному реагируют на различные типы медиатора. В [4] обосновывается, что синапс может выступать как сложная вычислительная система, состоящая из множества логических элементов.

Противоречие заключается в следующем:

– с точки зрения физики, весовой коэффициент синапса определяется не столько его размером, а сколько близостью его расположения к основному телу (соме) нейрона. То есть, чем ближе синапс находиться к центру нейрона, тем он сильнее воздействует на возбуждение нейрона и наоборот;

– чем больше размер синапса, тем больше рецепторов он имеет и тем более сложной вычислительной системой он является. Какую логическую функцию может реализовывать синапс, если с точки зрения нейросети он имеет фактически один вход и один выход?

Авторы статьи предлагают следующую интерпретацию приведённых фактов – каждый синапс запоминает, как необходимо реагировать на *различные* интенсивности входных воздействий. То есть, по мере обучения нейрона, каждый его синапс запоминает, насколько сильно он должен стимулировать или подавлять возбуждение нейрона для каждой конкретной интенсивности входного воздействия. Соответственно, чем больше опыта получает нейрон, тем более развитым должен быть синапс, чтобы помнить пары «интенсивность входного воздействия на синапс – интенсивность воздействия синапса на нейрон»

Эта интерпретация значительно отличается от общепринятой, в которой считается, что чем сильнее интенсивность входного воздействия на синапс, тем сильнее синапс стимулирует или подавляет возбуждение нейрона.

С точки зрения математической модели искусственного нейрона предлагается заменить операцию умножения входного значения и весового коэффициента на операцию табличной подстановки.

**Общеизвестная модель искусственного нейрона:**

Для лучшего понимания предлагаемого подхода рассмотрим модель формального нейрона [2]

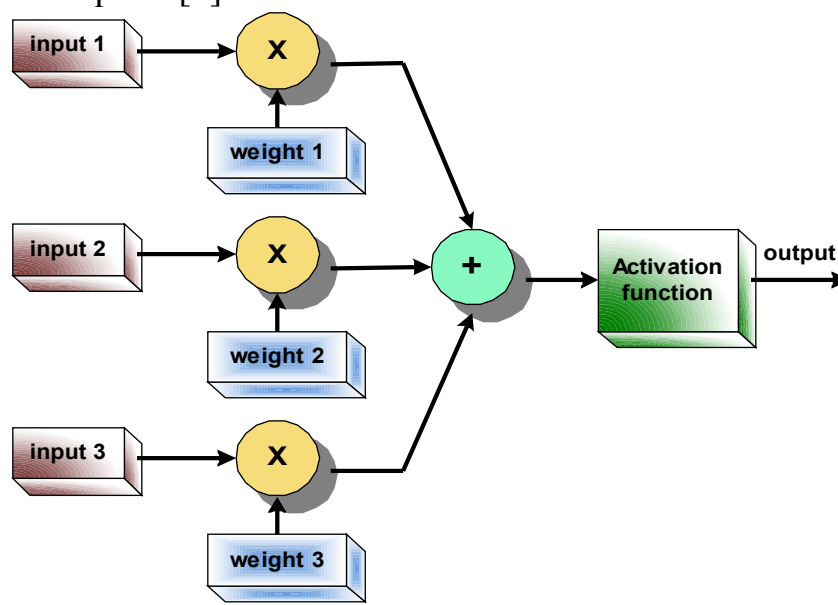

***Рисунок 3.** Функциональная схема формального нейрона (случай с тремя входами)*

Формальный нейрон представляет собой математическую модель простого процессора, имеющего несколько входов и один выход. Вектор входных сигналов (поступающих через "дендриды") преобразуется нейроном в выходной сигнал (распространяющийся по "аксону") с использованием трех функциональных блоков: блока перемножения входов с весовыми коэффициентами, блока суммирования и блока нелинейного преобразования (функция активации). Выбором весов достигается та или иная интегральная функция нейрона.

В блоке суммирования происходит накопление общего входного сигнала (обычно обозначаемого символом net), равного взвешенной сумме входов:

$$net = \sum_{i=1}^{N} weight_i \cdot input_i$$

В модели отсутствуют временные задержки входных сигналов, поэтому значение net определяет полное внешнее возбуждение, воспринятое нейроном. Отклик нейрона далее описывается следующим образом:

$$output = f(net)$$

где $f()$ – функция активации. В простейшем случае представляет из себя пороговую функцию, при этом $output=0$, если $net<\Theta$, и $output=1$, если $net\geq\Theta$ ($\Theta$ – значение порога). Может также использоваться линейная функция, сигмоидальная функция и т.д.

Следует отметить, что и в других известных моделях искусственного нейрона (модель Паде, квадратичный нейрон и т.д.) синаптическая связь также представляется как умножение входа на весовой коэффициент.

**Модель кибернейрона, предлагаемая авторами**, приведена на рисунке 4. Данная модель состоит из двух блоков: блока табличной подстановки и блока суммирования.

Каждое входное значение подаётся на соответствующую таблицу подстановки (sbox). Дискретное значение входа интерпретируется как индекс ячейки таблицы, а выходом является значение, хранящееся в ячейке. Выходы со всех таблиц подстановок суммируются, в результате формируется выход нейрона.

$$output = \sum_{i=1}^{N} sbox_i[input_i]$$

Модель имеет следующие особенности:
– Таблица подстановки соответствует синапсу биологического нейрона, а операция суммирования – суммарному воздействию синапсов на возбуждение нейрона. В соответствии с данной моделью каждый синапс, в зависимости от входного воздействия, может быть как тормозящим, так и возбуждающим (каждая ячейка таблицы может хранить как отрицательные, так и положительные числа).
– Размерность таблицы подстановки (количество ячеек таблицы) соответствует диапазону входных значений.

– Размерность хранящихся в ячейках таблицы чисел может быть произвольной и зависит от решаемой задачи.

– В данной модели отсутствует выходная нелинейная функция (функция активации). Это объясняется следующим: таблицей подстановки можно описать любую дискретную функцию, в том числе и функцию умножения дискретных чисел, и дискретные функции активации. Поэтому при помощи данной модели можно также реализовать формальный нейрон, использующий дискретные вычисления.

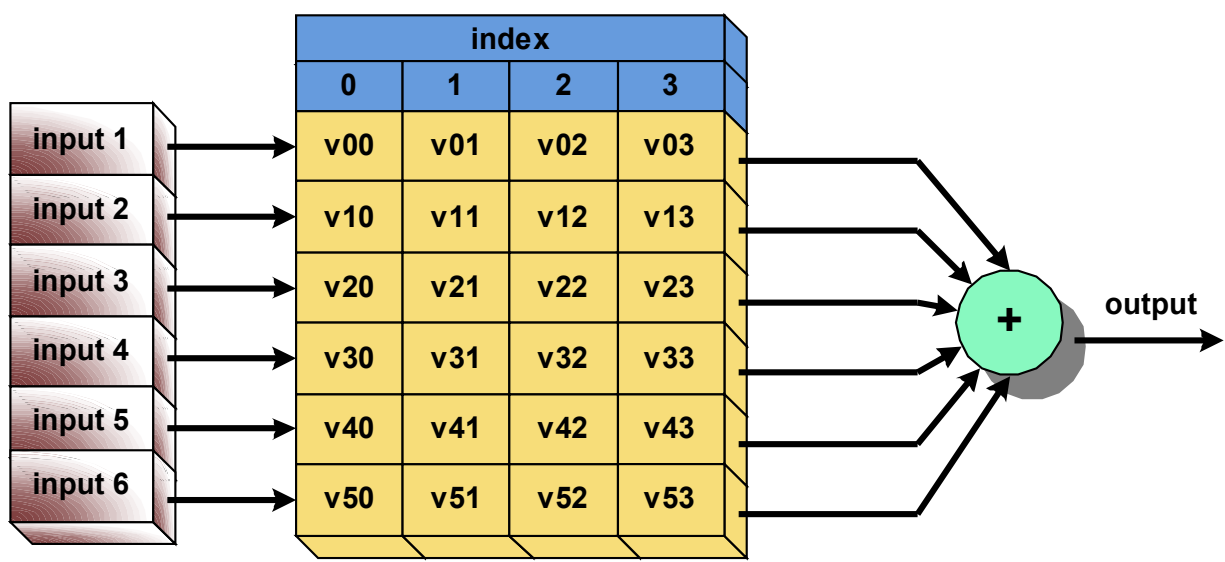

*Рисунок 4.* Функциональная схема кибернейрона (случай с шестью входами, размерность таблицы подстановки – 4)

Пример кибернейрона, обученного распознаванию образа (1,3,0,1,2,1) приведён на рисунке 6.

**Обучение кибернейрона.**

Введём следующие обозначения:
- input – вход кибернейрона;
- sbox – таблица подстановки, ассоциируемая с входом нейрона;
- output – выход кибернейрона;
- threshold – порог для выхода, при котором считается, что кибернейрон обучен образу (например, threshold = 100);
- threshold2 – порог для выхода, при котором считается, что кибернейрон не обучен образу (например, threshold2 = 20);
- N – количество входов кибернейрона;
- m – размерность таблицы подстановки (например, для 8-и битовых входов m = 256 и т.д.);
- k – коэффициент, отвечающий за скорость обучения;
- -- – операция декремента (вычитание единицы из значения);
- ++ – операция инкремента (добавление единицы к значению);

Целью обучения кибернейрона является такая модификация значений ячеек таблиц подстановок, при которых:
1. На выходе формируется значение output > threshold, если на вход подаётся ранее запомненный образ.
2. На выходе формируется значение output > threshold2 и output < threshold, если на вход подаётся новый образ, частично совпадающий с ранее запомненным образом.
3. На выходе формируется значение output < threshold2, если на вход подаётся новый образ, который не совпадает с ранее запомненными образами.

Таким образом, получается шкала, показывающая, насколько поданный на вход образ похож с ранее запомненными нейроном образами (рисунок 5). Указанные значения порогов приведены для примера, точные значения порогов в общем случае зависят от решаемой задачи.

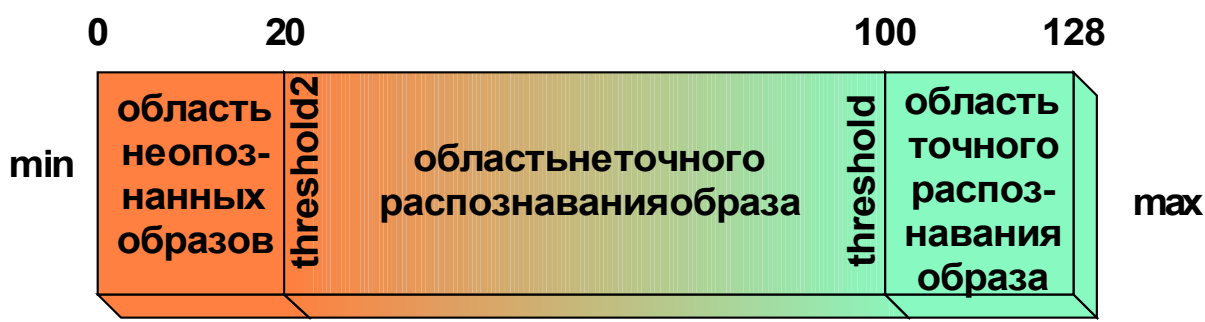

*Рисунок 5. Пример шкалы соответствия*

В общем случае возникает две задачи обучения кибернейрона – задача добавления (запоминания) нового образа и задача удаления образа из памяти. Вторая задача особенно важна при распределении большого количества образов между множеством параллельно включенных нейронов. На основании всего вышесказанного, авторами было разработано несколько алгоритмов обучения кибернейрона.

**1. Алгоритм простого обучения, использующий последовательную модификацию отдельных ячеек таблиц подстановок.** Для обучения кибернейрона новому образу необходимо выполнить следующие шаги:
1) для подаваемого на вход образа вычисляется выходное значение кибернейрона;
2) если выход больше верхнего порога, то завершаем процедуру бучения;
3) иначе, вычисляем разницу между выходом и порогом; полученное значение показывает, сколько раз требуется инкрементировать значения активных ячеек;
4) последовательно перебираем таблицы подстановок (от 0 до N-1) и увеличиваем на «1» значения ячеек, участвовавших в формировании

выхода (до полной компенсации разницы между выходом и порогом);

При необходимости добавления других образов, для них также осуществляются шаги 1–4, после чего производится проверка правильности распознавания образов. Если возникают несоответствия, то процедура обучения повторяется.

Удаление образа производиться аналогично процедуре добавления образа, только вычисляется разница между нижним порогом и выходом кибернейрона и путём уменьшения значений активных ячеек осуществляется компенсация разницы.

**2. Алгоритм простого обучения, использующий псевдослучайную модификацию отдельных ячеек таблиц подстановок.** Данный алгоритм отличается от предыдущего тем, что вместо последовательного перебора модифицируемых ячеек используется случайный перебор ячеек таблиц подстановок. Это позволяет внести элемент случайности в процесс обучения кибернейрона – при одинаковых обучающих выборках нейроны будут обучаться по-разному, т.е. по-разному фиксировать характерные черты образов. В ряде практических задач данная особенность может быть очень востребована (например, при распознавании модификаций компьютерных вирусов).

В соответствии с приведённым алгоритмом была разработана программная реализация процесса обучения кибернейрона. Входными данными являются: teaching_data – флаг обучения, если равен 1, то требуется добавить образ, если равен 0 – удалить образ; sbox – набор таблиц подстановок, составляющих кибернейрон. Перед процессом обучения должен быть запущен процесс определения выхода кибернейрона, в результате чего нам будут известны выход нейрона и ячейки таблиц, участвующих в формировании выхода.

Пример функции обучения кибернейрона на языке С++:

```cpp
void Educate_neuron(int teaching_data, BYTE **sbox)
{
 int threshold = 100;                       // верхний порог;
 int threshold2 = 20;                       // нижний порог;
 int divider = 4;                           // divider = 1/(коэфф. обучения)
 int modifier;                              // разница между порогом и вы-
                                            //   ходным значением нейрона

 if(teaching_data == 1)                     // если требуется обучить ней-
 {                                          //   рон образу,
    modifier = threshold - output;          // то вычисляем разницу между
      modifier = modifier/divider;          //   верхним порогом и выходом
     if(modifier == 0) {modifier = 1;}      //   нейрона
     ChangeNeuronStats(modifier, sbox);     // и модифицируем параметры
                                            //   нейрона;
 }

 else                                       // если требуется, чтобы  нейрон
```

```
        {
            modifier = threshold2 - output;          // то вычисляем разницу между
            modifier = modifier/divider;             нижним порогом и выходом
            if(modifier == 0) {modifier = -1;}       нейрона
            ChangeNeuronStats(modifier, sbox);       // и модифицируем параметры
        }                                            нейрона;
    }
}
```
                                                     не распознавал образ,

Пример функция изменения параметров кибернейрона на языке С++:

```
void ChangeNeuronStats(int modifier, BYTE **sbox)
{
  int row_number;                                     //номер таблицы подстановки;
  int cell_number;                                    //номер ячейки в таблице;

if(modifier >=0 )                                     //если разница больше 0,
  for (int m = 0; m < modifier; m++)                  //то выполняем цикл, пока раз-
    {                                                  ница не будет компенсирована
     row_number = random(N);                          //выбираем таблицу;
     cell_number = activating_cells_in_sbox[row_number]; //определяем для неё активную
                                                         ячейку;
       if(sbox[row_number][cell_number] < 127)
       { sbox[row_number][cell_number]++;}            //увеличиваем значение ячейки
    }                                                 на 1;

  else                                                //если разница меньше 0,
  for (int m = 0; m < abs(modifier); m++)             //то выполняем цикл, пока раз-
    {                                                  ница не будет компенсирована
     row_number = random(N);                          //выбираем таблицу;
     cell_number = activating_cells_in_sbox[row_number]; //определяем для неё активную
                                                         ячейку;
       if(sbox[row_number][cell_number] > -126)
       {sbox[row_number][cell_number]--; }            //уменьшаем значение ячейки
    }                                                 на 1;
}
```

Следует отметить, что приведённые алгоритмы обучения не являются единственными и наиболее оптимальными. Они приведены для того, чтобы показать принципиальное отличие процесса обучения кибернейрона от процесса обучения формального нейрона. Данное отличие в процессах обучения объясняет преимущества кибернейрона.

В случае обучении нейрона новому образу, вероятность искажения предыдущих запомненных образов снижается пропорционально размерности таблиц подстановок. Эта вероятность *значительно* ниже, по сравнению с формальным нейроном. Объясняется тем, что при обучении модифицируется только одна ячейка из каждой таблиц подстановок. В формальном нейроне при обучении меняются весовые коэффициенты входов, что сразу влияет на *все* запомненные образы.

Количество запоминаемых образов кибернейрона зависит от размерности таблиц подстановок, а также от количества входов. С увеличением этих параметров растёт информационная ёмкость кибернейрона (рисунки 10,11). Сам процесс распознавания образов при этом напоминает кодовое разделение сигналов (CDMA).

Для более простого понимания предлагаемых идей приведём ряд примеров. На рисунке 6 изображён кибернейрон, имеющий шесть входов. Каждому входу соответствует таблица подстановки размерностью 4 ячеек. Кибернейрон обучен распознаванию одного образа – (1,3,0,1,2,1).

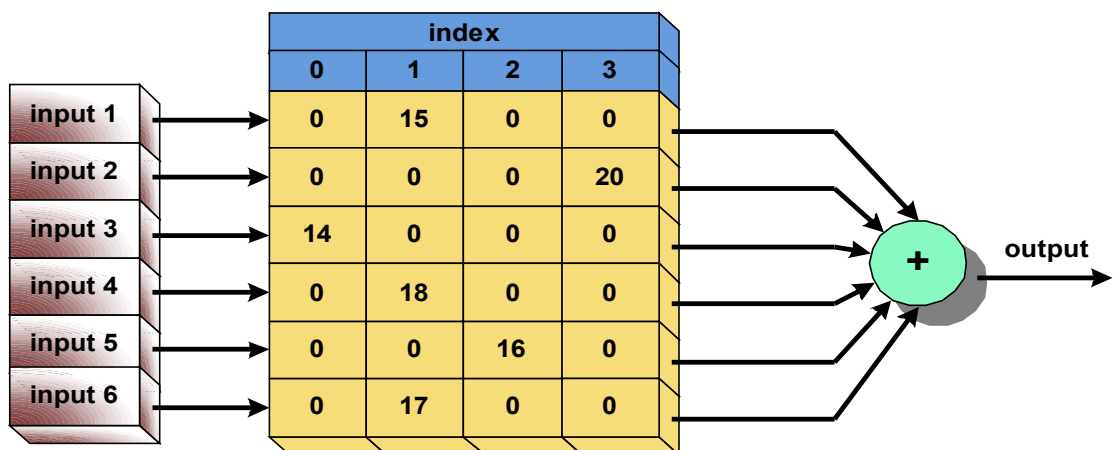

**Рисунок 6.** *Пример кибернейрона, обученного образу (1,3,0,1,2,1)*

На рисунке 7 изображён процесс распознавания кибернейроном этого же образа. Другим цветом отмечены активные ячейки. А на рисунке 8 изображён процесс распознавания образа, который незначительно отличается от обученного образа. При этом видно, что две таблицы подстановки дали на своём выходе значение 0. Однако суммарное значение на выходе данной модели достаточно высокое (68 из 100), что показывает достоверность работы модели.

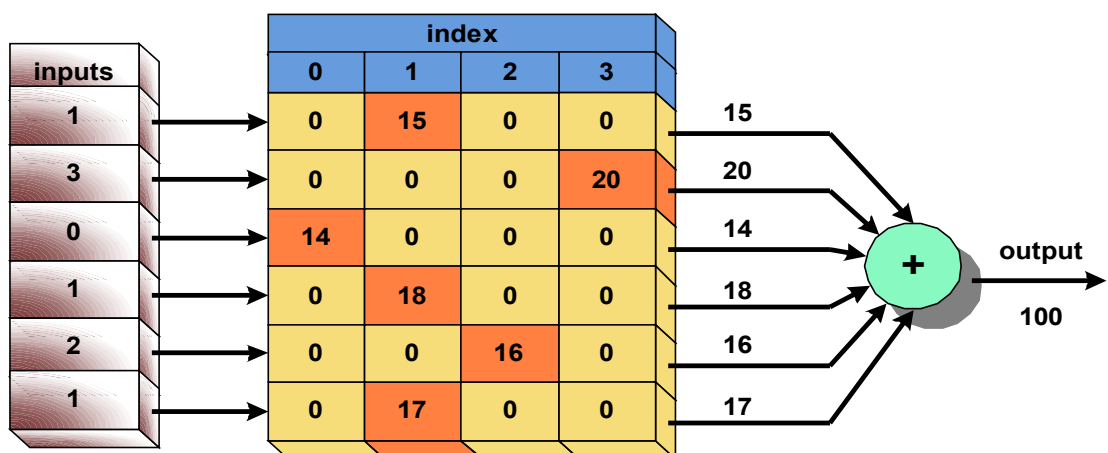

**Рисунок 7.** *Пример работы кибернейрона, при подаче на его вход образа (1,3,0,1,2,1)*

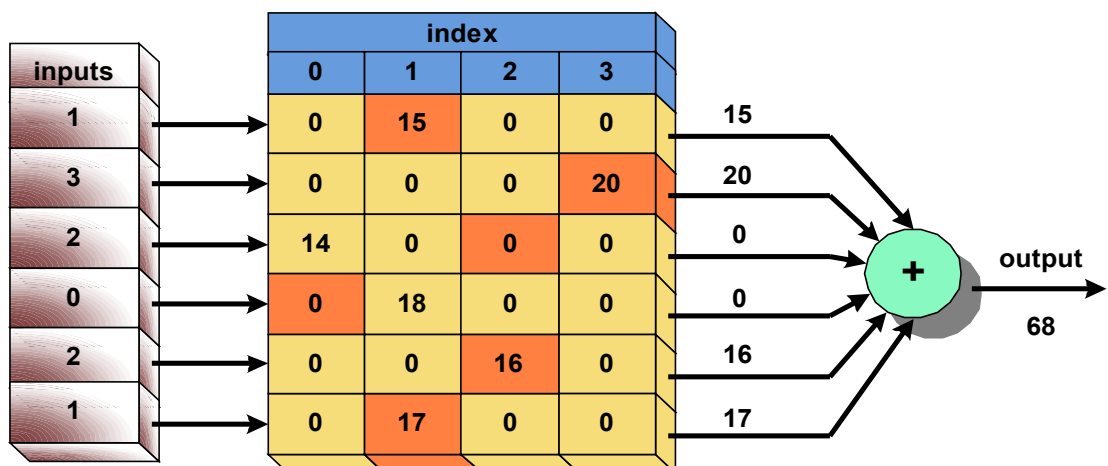

**Рисунок 8.** *Пример работы кибернейрона, при подаче на его вход образа (1,3,2,0,2,1)*

На рисунке 9 показан пример кибернейрона, имеющего 8 входов по 256 элементов каждый (кибернейрон обучен 200-м образам). Ячейкам, хранящих нулевые значения, соответствует серый цвет. Ячейкам, хранящих положительные значения, соответствуют светлые оттенки, а хранящим отрицательные значения – тёмные оттенки серого.

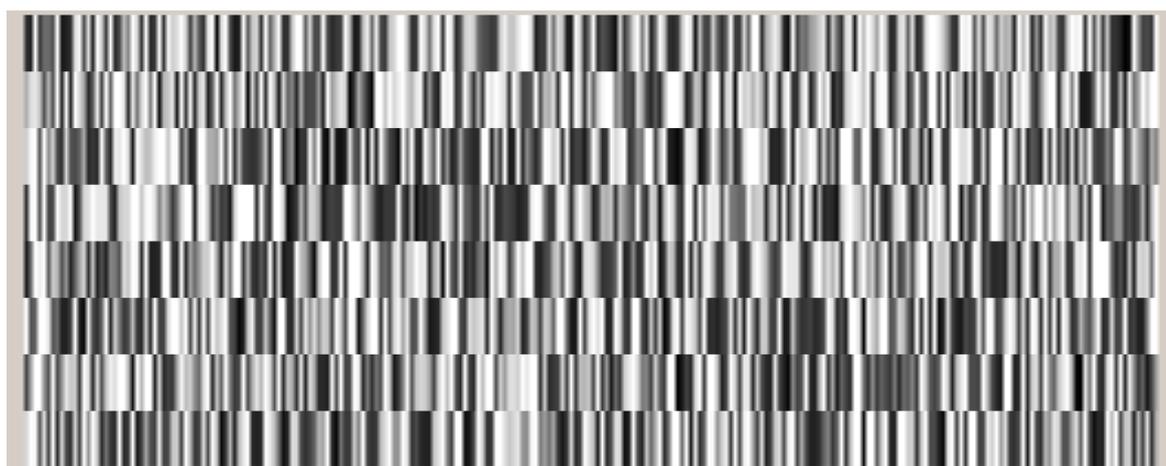

**Рисунок 9.** *Пример обученного кибернейрона, имеющего 8 входов по 256 элементов каждый (кибернейрон обучался 200-м образам)*

Для более детальной оценки эффективности предложенной модели искусственного нейрона были проведены экспериментальные исследования свойств кибернейрона. Для этого исследовались:
- зависимость доли правильно обученных образов от количества итераций обучения;
- зависимость доли ошибочного распознавания образов от количества итераций обучения;
- зависимость количества запоминаемых образов от коэффициента обучения.

Исследовались кибернейроны, имеющие 8-и битные и 16-и битные входы. Образы для обучения формировались случайным образом. Результаты исследования, для уменьшения занимаемого объема, сведены в графики.

**Характеристики кибернейрона, имеющего 8-и битные входы (256 ячеек для каждой таблицы подстановки).** Использовался коэффициент обучения k = 0,25.

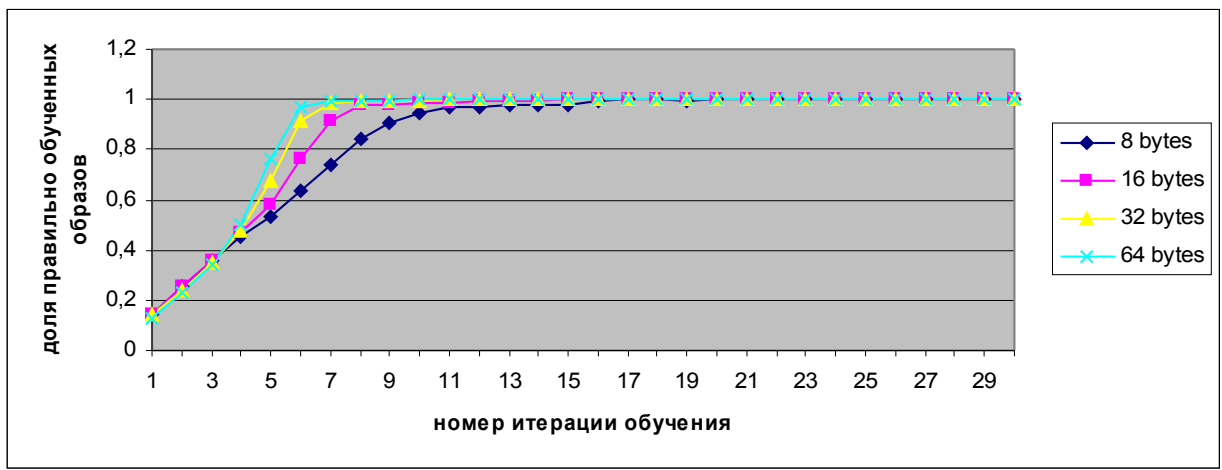

*Рисунок 10.* *Зависимость доли правильно обученных образов от количества итераций обучения (использовалось 200 образов, данные приведены для обучения 8-и,16-и,32-х и 64-х байтным образам)*

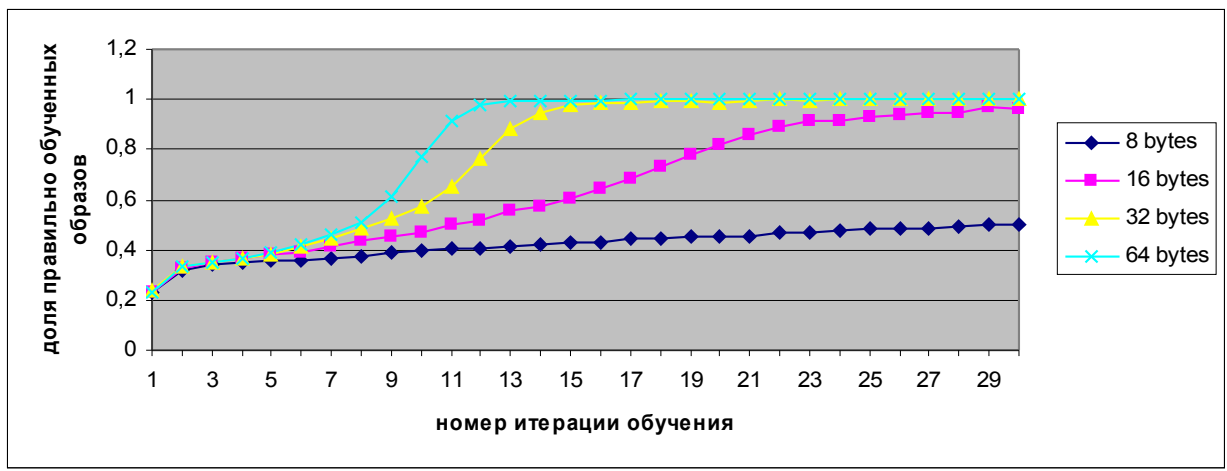

*Рисунок 11.* *Зависимость доли правильно обученных образов от количества итераций обучения (использовалось 800 образов, данные приведены для обучения 8-и,16-и,32-х и 64-х байтным образам)*

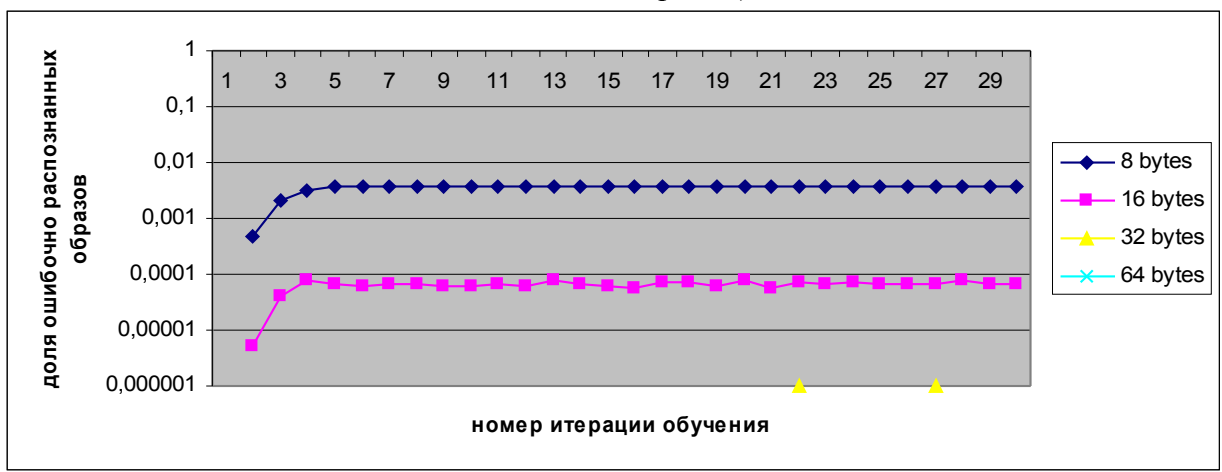

*Рисунок 12.* *Зависимость доли ошибочного распознавания образов от количества итераций обучения (использовалось 200 образов, данные приведены для обучения 8-и,16-и, 32-х и 64-х байтным образам)*

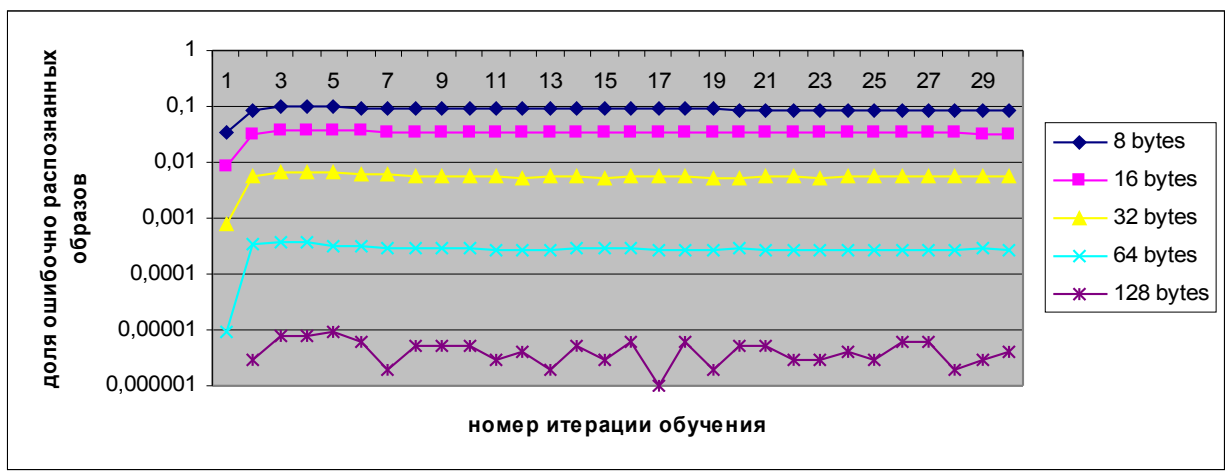

*Рисунок 13.* Зависимость доли ошибочного распознавания образов от количества итераций обучения (использовалось 800 образов, данные приведены для обучения 8-и,16-и, 32-х и 64-х байтным образам)

**Характеристики кибернейрона, имеющего 16-и битные входы (65536 ячеек для каждой таблицы подстановки).** Использовался коэффициент обучения k = 0,25.

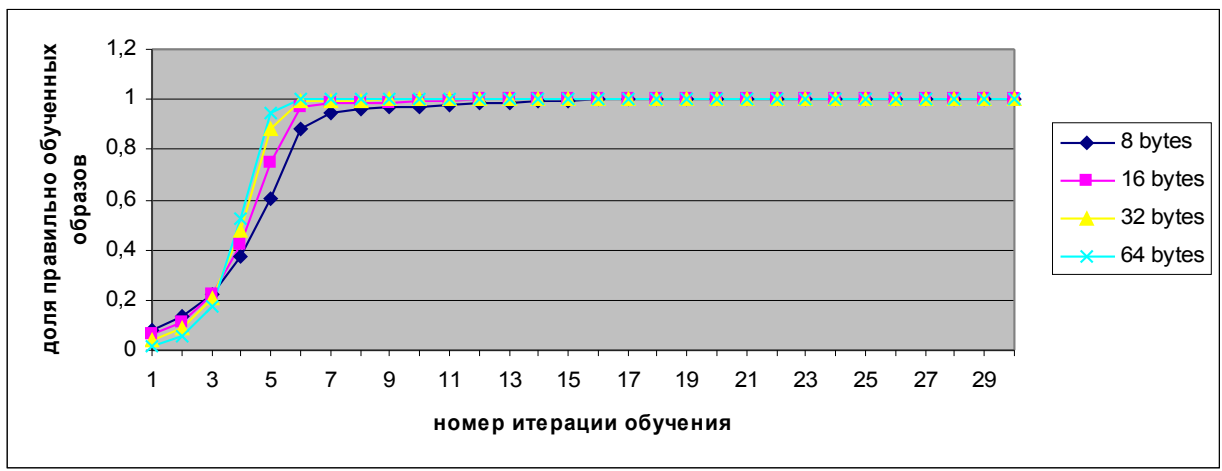

*Рисунок 14.* Зависимость доли правильно обученных образов от количества итераций обучения (использовалось 16384 образов, данные приведены для обучения 8-и,16-и,32-х и 64-х байтным образам)

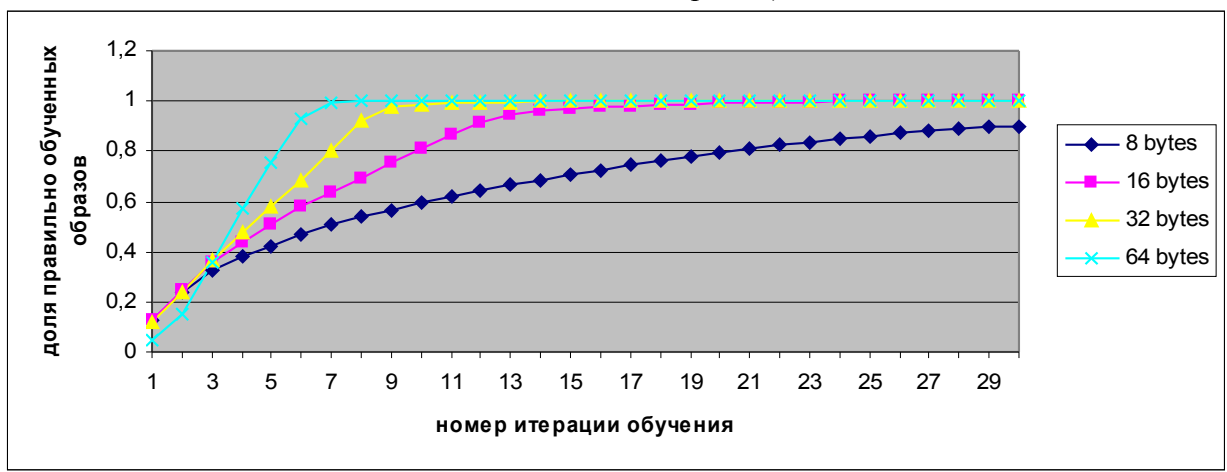

*Рисунок 15.* Зависимость доли правильно обученных образов от количества итераций обучения (использовалось 65536 образов, данные приведены для обучения 8-и,16-и,32-х и 64-х байтным образам)

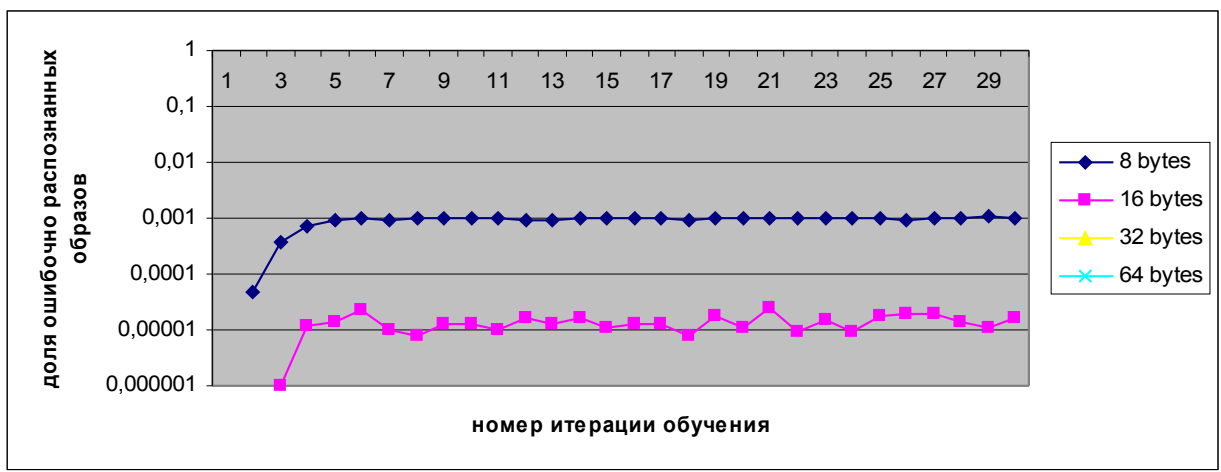

*Рисунок 16.* Зависимость доли ошибочного распознавания образов от количества итераций обучения (использовалось 16384 образов, данные приведены для обучения 8-и, 16-и,32-х и 64-х байтным образам)

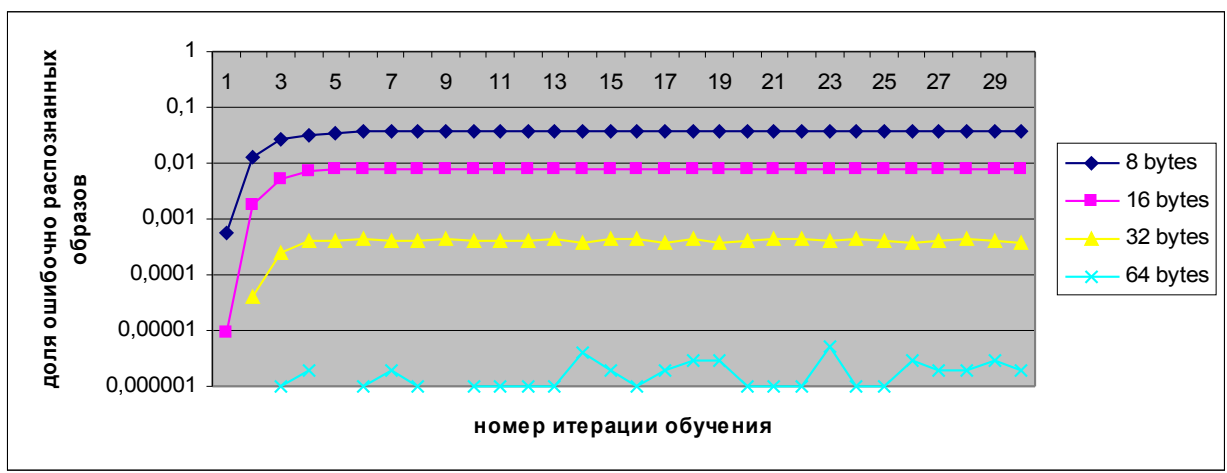

*Рисунок 17.* Зависимость доли ошибочного распознавания образов от количества итераций обучения (использовалось 65536 образов, данные приведены для обучения 8-и, 16-и,32-х и 64-х байтным образам)

**Влияние коэффициента обучения на характеристики кибернейрона, имеющего 8-и битные входы, (256 ячеек для каждой таблицы подстановки).**

Целью данного экспериментального исследования являлось определение влияния коэффициента обучения на скорость обучения и на вероятность ошибочного распознавания образов. Результаты исследования показывают, что точность распознавания образов с уменьшением коэффициента увеличивается незначительно, при этом количество итераций обучения резко увеличивается. Поэтому, использование коэффициента обучения меньше 0,25 нежелательно.

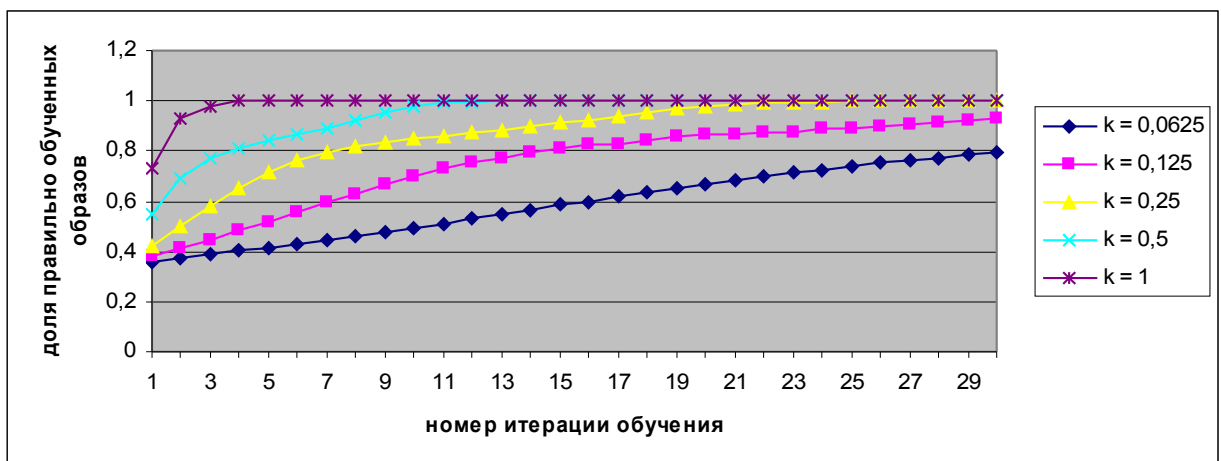

***Рисунок 18.*** *Влияние коэффициента обучения на скорость обучения (использовалось 400 образов, данные приведены для обучения 8-и байтным образам)*

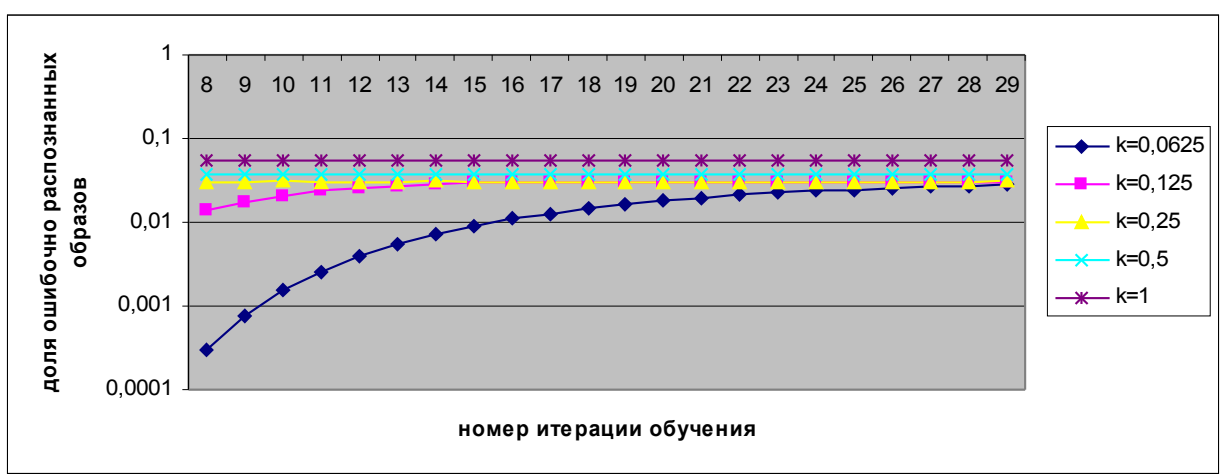

***Рисунок 19.*** *Влияние коэффициента обучения на вероятность ошибочного распознавания образов (использовалось 400 образов, данные приведены для обучения 8-и байтным образам)*

Приведённые экспериментальные данные говорят о достаточно высокой информационной ёмкости одного кибернейрона. Информационная ёмкость крайне сильно зависит от количества входов и от размерности таблиц подстановок. Для увеличения информационной ёмкости достаточно увеличить размерности используемых таблиц подстановок. Также можно осуществлять размен «информационная ёмкость – скорость обучения». Всё это свидетельствует о высокой гибкости предложенной модели искусственного нейрона.

**Использование кибернейрона для сигнатурного поиска компьютерных вирусов.**

Для проверки эффективности предложенной модели искусственного нейрона использовался открытый антивирус ClamAV [7]. Данный антивирус имеет статус opensource проекта, распространяется по лицензии GPLv2. Проект начал функционировать в 2002 году. Данный антивирус является фактически единственным эффективным антивирусом с открытым исходным кодом, составляющий конкуренцию коммерческим антивирусам.

На март 2008 г. было осуществлено более двух миллионов загрузок ClamAV с крупного opensource ресурса www.sourceforge.org (переводится дословно как «кузница исходных кодов»). Здесь не учитывается количество загрузок с официального сайта ClamAV (www.clamav.com).

**Особенности ClamAV**. Основное назначение ClamAV – сканирование файлов на почтовых серверах и т.д. Содержит более 169.676 сигнатур вредоносного кода. Не позволяет лечить файлы – заражённые файлы либо удаляются, либо помещаются в карантин. Использует сигнатурный метод поиска компьютерных вирусов. Упрощённа структура ClamAV изображена на рисунке 20.

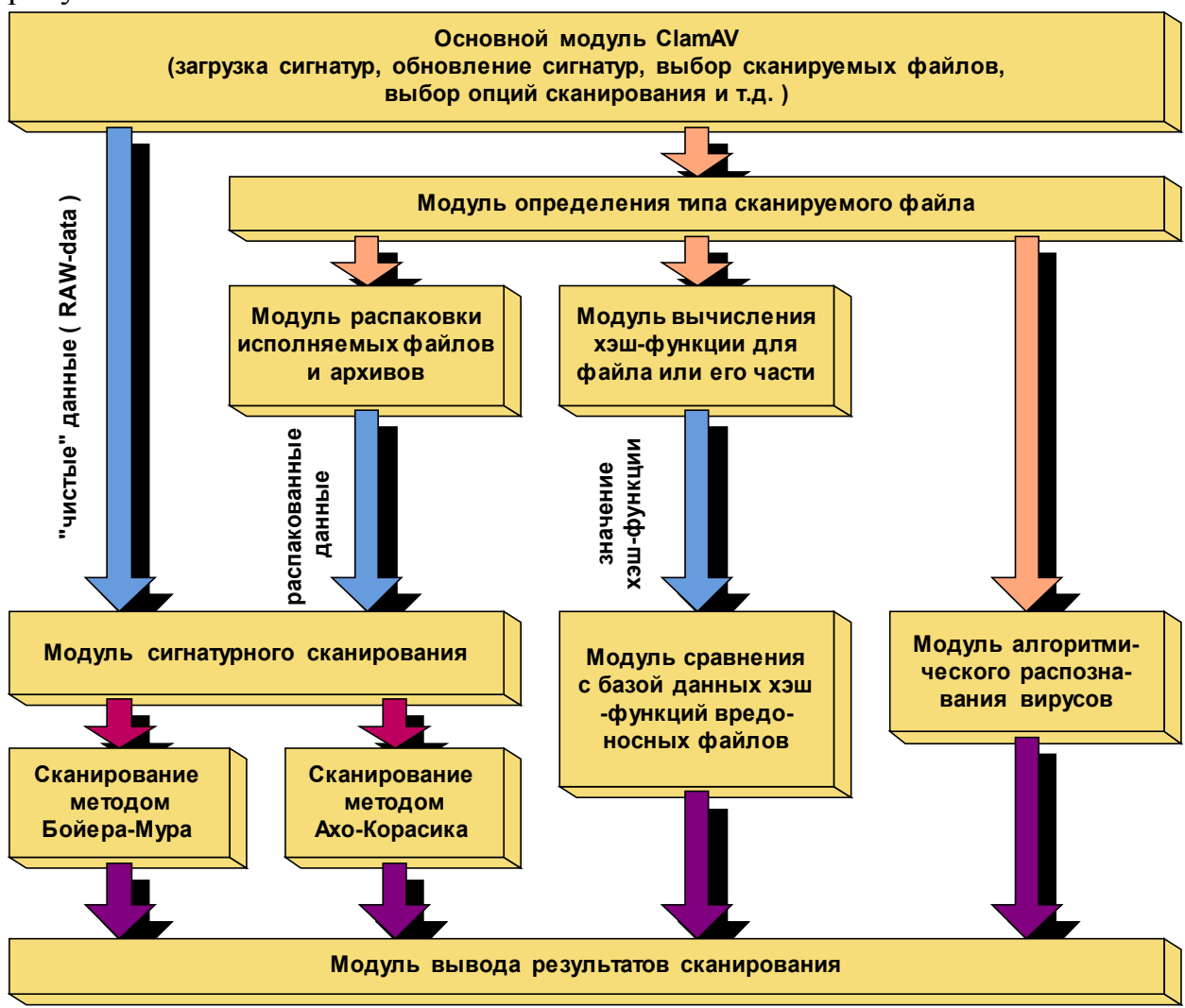

***Рисунок 20.*** *Упрощённая структура антивируса ClamAV*

ClamAV использует несколько вариантов описания сигнатур вирусов:
- Описание сигнатуры в виде последовательности шестнадцатеричных цифр. Данный тип сигнатур хранится в файле с расширением *.db. Пример сигнатуры:
  «Phantom.4=0190e800005e56ba4c0881ea000183ee».

Допускается использование масок, т.е. замена шестнадцатеричных значений символами «*», «?» и т.д.
- Сигнатуры описываются так же, как и в предыдущем случае. Дополнительно указывается номер и тип секции исполняемого файла, в котором может находиться вредоносный код. Дополнительные сведения служат для увеличения скорости работы антивируса. Данный тип сигнатур хранится в файле с расширением *.ndb. Пример сигнатуры:

    «W32.MyLife.E:1:*:7a6172793230*40656d61696c2e636f6d»
- Сигнатура представляется в виде результата вычисления хэш-функции от файла, содержащего вредоносный код. Данный вариант применим в случае, когда вирус размещается в отдельном файле. Неприменим для полиморфных и самошифрующихся вирусов, а так же для вирусов, заражающих другие файлы. Достоинство – простота получения сигнатуры для вируса.

    Для ускорения работы антивируса дополнительно указывается размер файла, содержащий вредоносный код. Это позволяет исключить вычисление хэш-функций для файлов, размер которых не соответствует размерам, приведённым в базе сигнатур. Данный тип сигнатур хранится в файлах с расширением *.mdb и *.hdb. Пример сигнатуры:

    «36864:d1a320843e3a92fdbb7d49137f9328a0:Trojan.Agent-1701»

**Быстродействие ClamAV**. В ClamAV применяется три основных метода поиска сигнатур в файлах: метод Бойра-Мура [8], метод Ахо-Корасика [9] и сравнение результатов хэш-функции. Следует отметить, что со времени запуска проекта ClamAV (2002 год) программные реализации методов Бойра-Мура и Ахо-Корасика постоянно совершенствовались, особенно в плане быстродействия.

Для оценки эффективности этих методов была определена скорость сканирования файлов антивирусом ClamAV на компьютере, имеющего следующую конфигурацию:
- процессор – AMD Athlon(tm) 64 X2 Dual Core;
- рабочая частота каждого ядра процессора – 2 ГГц;
- оперативная память – DDR2, 667 МГц, объёмом 3 Гб;
- жёсткий диск – Samsung SP2514N, 250 Гб;
- операционная система – Windows XP.

Скорость оценивалась по времени обработки файла размером 615 Мбайт, состоящего из смеси исполняемых файлов, библиотек dll и т.д. Режим сканирования – RAW (т.е. сканирование чистых данных, без распаковки исполняемых файлов и архивов).

Скорость сканирования определялась несколько раз и составила **6,7 Мбайт** в секунду. После первого замера, тестовый файл полностью размещался в кэш памяти ОС Windows, что устраняло влияние скорости считывания жёсткого диска на скорость сканирования.

**Модификация ClamAV.** Для увеличения быстродействия антивируса ClamAV в него был добавлен модуль быстрого детектирования вредоносного кода в сканируемых данных (рисунок 21). Данный модуль представляет собой программную реализацию кибернейрона, имеющего 2 входа по 24 бит каждый. Соответственно размерность таблицы подстановки для каждого входа составила 16.777.216 ячеек по 8 бит каждая. Всего под данный кибернейрон было отведено дополнительно 32 Мбайт оперативной памяти.

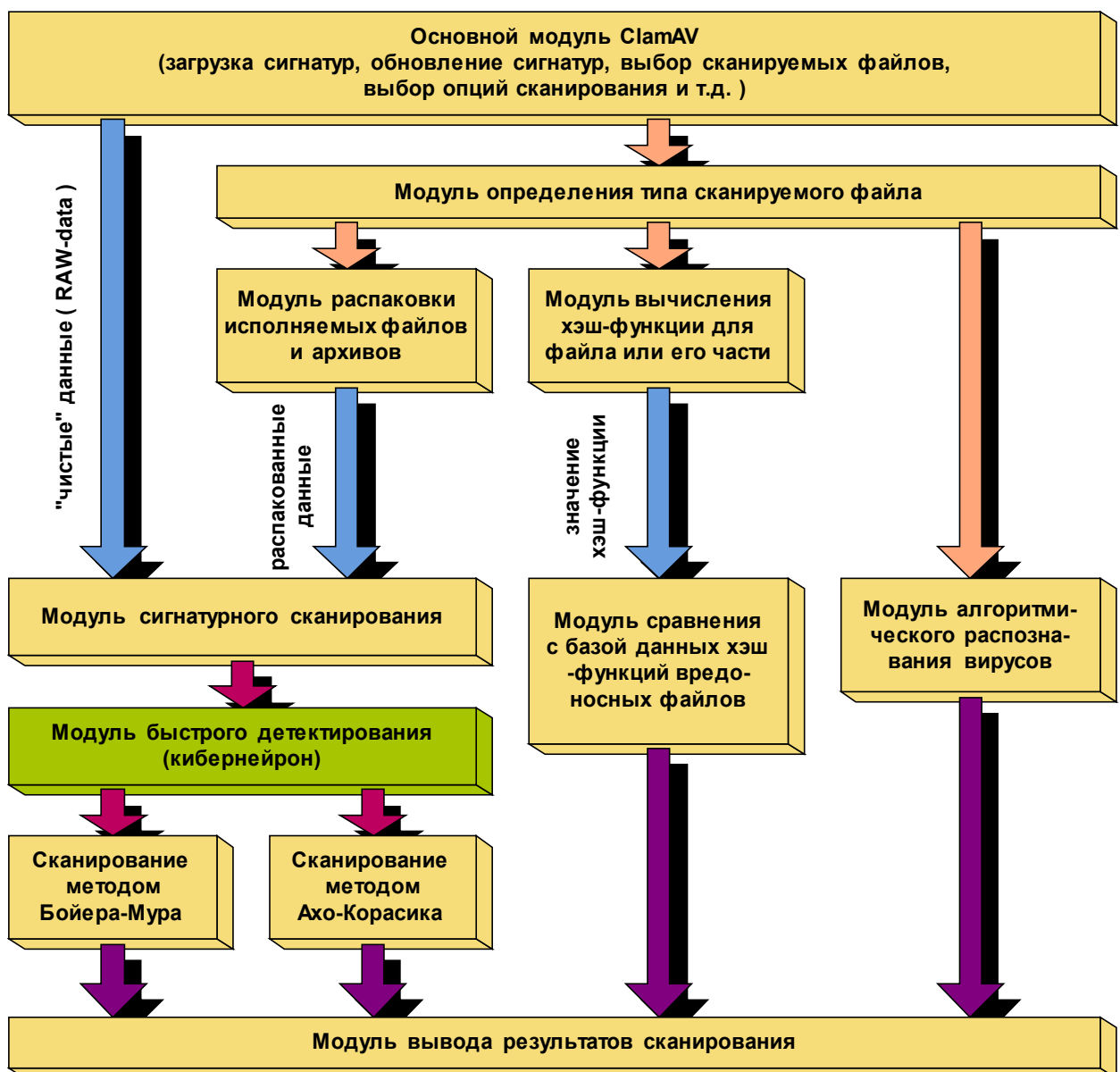

***Рисунок 21.*** *Добавление в ClamAV дополнительного блока быстрого детектирования*

Сканируемые данные разбивались на неперекрывающиеся блоки длиной по 6 байт каждый, что соответствует размерности входа кибернейрона (2 по 24 бит). Задачей кибернейрона являлась выдача результата: 1 – обработанный блок информации содержит часть сигнатуры; 0 – не содержит. В случае, если кибернейрон обнаруживал совпадение, то блок информации и соседние с ним блоки передавались стандартным средствам обнаружения ClamAV.

Для предотвращения пропуска участка, совпадающего с сигнатурой вируса, кибернейрон обучался не только 6-и байтному участку сигнатуры, но и 6-и байтным участкам, полученных путём сдвига на 1,2,3,4 и 5 позиций (см. рисунок 22). Таким образом, каждой исходной сигнатуре соответствовало шесть 6-и байтных сигнатур.

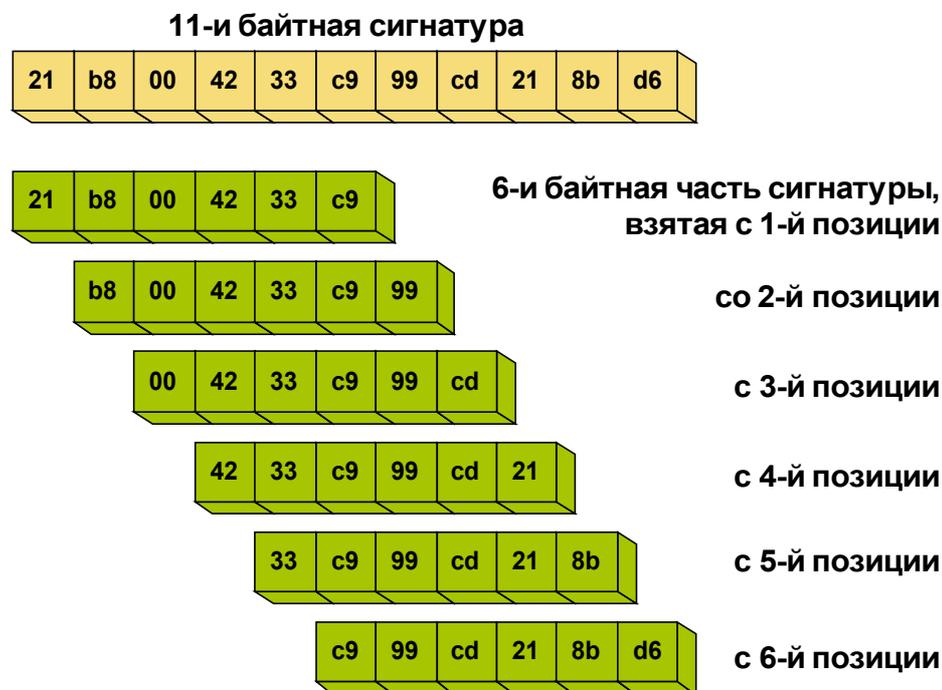

***Рисунок 22.*** *Принцип формирования обучающих выборок для блока быстрого детектирования*

Для обучения кибернейрона использовались сигнатуры из файлов *.db и *.ndb. При этом были отсеяны сигнатуры, имеющие длину менее 11 байт, а также сигнатуры, содержащие в себе куски, которые часто встречаются в незаражённых данных. Например, содержащие длинные последовательности нулей, содержащие данные заголовков исполняемых файлов и т.д. В процессе отсеивания было отброшено около 35 % сигнатур.

Такое внушительное количество отброшенных сигнатур объясняется тем, что данные сигнатуры составлялись без учёта особенностей нейронного обнаружителя.

Всего из указанных файлов было загружено 114464 сигнатур, отсеяно 40179 сигнатур и извлечено 74288 11-и байтных сигнатур. Кибернейрон был обучен 445728 6-и байтным сигнатурам, полученных из 11-и байтных сигнатур (рисунок 22).

Скорость сканирования модифицированным ClamAV определялась несколько раз и составила ~ **22 Мбайт** в секунду. Корректность работы модифицированного ClamAV была проверена на коллекции вирусов, состоящей из 3721 файлов (обнаружено 3598 вирусов из 3721, или 96,6 %).

Дополнительно была определена скорость сканирования при использовании сигнатур, сформированных псевдослучайным образом. Скорость сканирования составила ~ **25 Мбайт** в секунду. При этом использовалось 100.000 11-и байтных сигнатур, или 600.000 6-и байтных сигнатур.

Таблица 1. Производительность ClamAV и его модификаций

| Антивирус | Скорость сканирования RAW данных, Мбайт/с |
|---|---|
| ClamAV | 6,7 |
| Модифицированный ClamAV | 22 |
| Модифицированный ClamAV + псевдослучайные сигнатуры | 25 |
| CyberDemon | 71 |

Приведённые результаты показывают эффективность использования кибернейрона для ускорения работы антивируса ClamAV. Единственной проблемой, мешающей полноценной интеграции кибернейрона в ClamAV, является проблема извлечения качественных 6-и байтных сигнатур из стандартной базы сигнатур ClamAV. Для решения этой проблемы требуется полная коллекция вирусов, что позволит выделить сигнатуры, наиболее оптимально подходящие для обучения кибернейрона.

**Антивирус CyberDemon.** Существенным фактором, ограничивающим скорость сканирования файлов при помощи модифицированного ClamAV, является поиск вредоносных файлов по хэш-функциям. Так же, кибернейрон использовался как элемент ускорения работы ClamAV, но не использовалась основная способность искусственного нейрона – способность обнаруживать модификации вирусных сигнатур.

Для определения эффективности применения кибернейрона для обнаружения вредоносных программ авторами был разработан собственный метод обнаружения вирусных сигнатур. На основе этого метода был разработан антивирус CyberDemon, упрощённая структурная схема которого представлена на рисунке 23.

Его отличительной особенностью является обнаружение вирусных сигнатур в 3 этапа:
1. Быстрое детектирование при помощи двух входового кибернейрона. Характеризуется крайне высокой скоростью сканирования данных (более 70 мегабайт в секунду) и достаточно высокой вероятностью ложных срабатываний (приблизительно 0.03%).
2. Точное обнаружение при помощи восьми входового кибернейрона. Характеризуется средней скоростью сканирования и очень низкой вероятностью ложного срабатывания (например, при сканировании файла длиной 800 мегабайт эта вероятность составила 0%).
3. В случае обнаружения в сканируемом блоке вредоносного кода, для него производится поиск наиболее близкого соответствия по базе вирусных сигнатур.

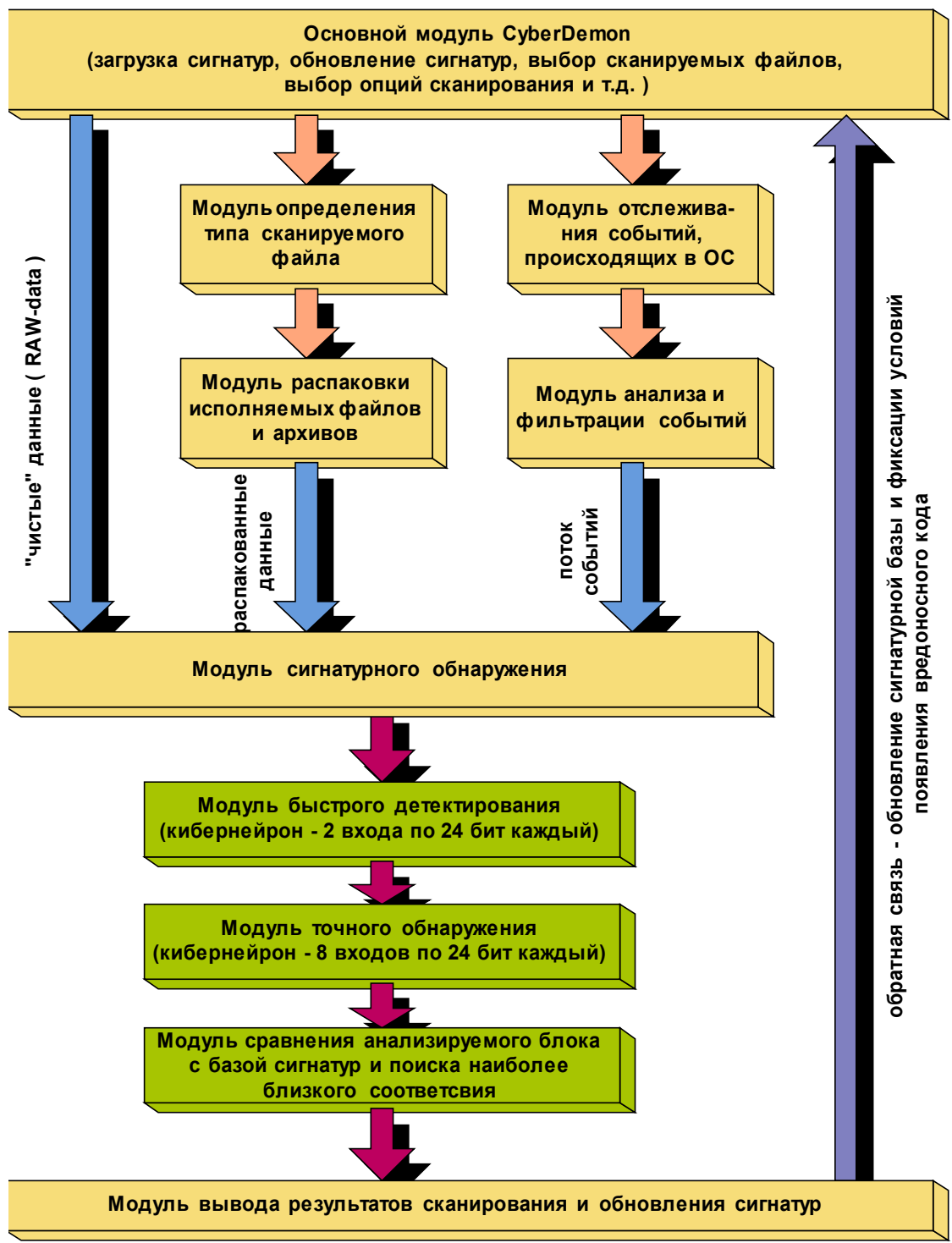

***Рисунок 23.*** *Упрощённая структура CyberDemon'а*

Использованием этих этапов достигается высокая скорость сканирования и возможность обнаружения ***модификаций*** вредоносного кода. Здесь может проявиться полезная особенность процесса обучения кибернейрона. Она заключается в том, что одинаковые нейроны по различному фиксируют характерные черты запоминаемых образов. Это позволяет создать распределенную антивирусную систему, состоящую из множества компьютеров, объединенных в единую сеть, каждый элемент которой обнаруживает различные характерные черты вредоносного кода и обменивается с другими эле-ментами информацией о новых модификациях вирусов. Данная система по эффективности обнаружения новых и модифицированных вирусов будет превосходить любую одиночную систему.

В силу рассмотренных в статье особенностей кибернейрона появляется возможность создания адаптивного антивируса, позволяющего как находить новые варианты вредоносного кода, так и обновлять базу вирусных сигнатур.

При существующих тенденциях увеличения количества компьютерных вирусов (2005 год – более 50 тысяч сигнатур; 2006 год – более 100 тысяч сигнатур; 2007 год – 220172 тысяч сигнатур [11-13]) может возникнуть ситуация, когда классические сигнатурные методы поиска компьютерных вирусов окажутся несостоятельными. Например, скорость работы антивируса окажется неудовлетворительной или потребуется необоснованно большое количество вычислительных ресурсов.

Для решения этой проблемы необходимо переходить к методам, обнаруживающих вредоносный код по результатам его деятельности – т.е., к сигнатурам событий. В этом случае в полной мере могут проявиться такие качества поиска при помощи кибернейрона, как способность обнаруживать похожие сигнатуры.

**Заключение:**
- Кибернейрон обладает значительно большей информационной ёмкостью, чем формальный нейрон.
- Обучение новым образам – быстрое, простое и требует только операций сложения и вычитания.
- При обучении кибернейрона новому образу, вероятность искажения предыдущих запомненных образов значительно ниже, по сравнению с формальным нейроном.
- Использование кибернейрона вместо формального нейрона в ряде случаев может значительно уменьшить сложность нейросети.
- Эффективная и простая реализация в программном и аппаратном видах.
- Вместо операции суммирования можно использовать другие операции объединения результатов.
- Экспериментальные исследования показали достаточно высокую эффективность применения кибернейрона при решении задач обнаружения компьютерных вирусов.
- Невозможность использования для обучения нейросети, составленной из кибернейронов, алгоритма обратного распространения ошибки
- В силу своей специфики использование кибернейрона в некоторых видах нейросетей может быть неэффективно (возникнет необходимость смены алгоритма обучения).

Проводимые авторами исследования в этом направлении не ограничиваются приведенным материалом. Уже получены принципы объединения кибернейронов в нейросеть, имеющую значительно большую информационную емкость и способную обучатся за 1 итерацию. Однако это является материалом отдельной статьи.